\pdfoutput=1

\documentclass[11pt]{article}

\usepackage{acl}
\usepackage{times}
\usepackage{latexsym}
\usepackage{multirow}
\usepackage{inconsolata}
\usepackage{booktabs}
\usepackage{colortbl}
\usepackage{rotating}

\usepackage{amsmath}
\usepackage{arydshln}
\usepackage{amsfonts}
\usepackage{xspace}
\usepackage{svg}
\PassOptionsToPackage{prologue,dvipsnames}{xcolor}
\newcommand{\model}{{\fontfamily{lmss}\selectfont MANGO}\xspace}
\newcommand{\w}{{\textbackslash}w}

\usepackage[T1]{fontenc}

\usepackage[utf8]{inputenc}

\usepackage{microtype}

\usepackage{inconsolata}
\usepackage{listings}
\usepackage{color}

\definecolor{dkgreen}{rgb}{0,0.6,0}
\definecolor{gray}{rgb}{0.5,0.5,0.5}
\definecolor{mauve}{rgb}{0.58,0,0.82}

\title{Comments as Natural Logic Pivots: Improve Code Generation via Comment Perspective}

\author{Yijie Chen$^1$, Yijin Liu$^2$, Fandong Meng$^2$, Yufeng Chen$^1$, Jinan Xu$^1$, Jie Zhou$^2$ \\
  $^1$Beijing Jiaotong University, Beijing, China\\
  $^2$Pattern Recognition Center, WeChat AI, Tencent Inc, China \\
    {\tt \{22120354, chenyf, jaxu\}@bjtu.edu.cn}\\
    {\tt \{yijinliu, fandongmeng, withtomzhou\}@tencent.com}
 }

\begin{document}
\maketitle
\begin{abstract}

Code generation aims to understand the problem description and generate corresponding code snippets, where existing works generally decompose such complex tasks into intermediate steps by prompting strategies, such as Chain-of-Thought and its variants. While these studies have achieved some success, their effectiveness is highly dependent on the capabilities of advanced Large Language Models (LLMs) such as GPT-4, particularly in terms of API calls, which significantly limits their practical applicability.
Consequently, how to enhance the code generation capabilities of small and medium-scale code LLMs without significantly increasing training costs is an appealing challenge.
In this paper, we suggest that code comments are the natural logic pivot between natural language and code language and propose using comments to boost the code generation ability of code LLMs. Concretely, we propose \model (comMents As Natural loGic pivOts), including a comment contrastive training strategy and a corresponding logical comment decoding strategy.
Experiments are performed on HumanEval and MBPP, utilizing StarCoder and WizardCoder as backbone models, and encompassing model parameter sizes between 3B and 7B.
The results indicate that \model significantly improves code pass rate based on the strong baselines. Meanwhile, the robustness of the logical comment decoding strategy is notably higher than the Chain-of-thoughts prompting. The code is publicly available at \url{https://github.com/pppa2019/Mango}.

\end{abstract}
\begin{figure}[ht]
    \centering
    \includegraphics[width=0.8\linewidth]{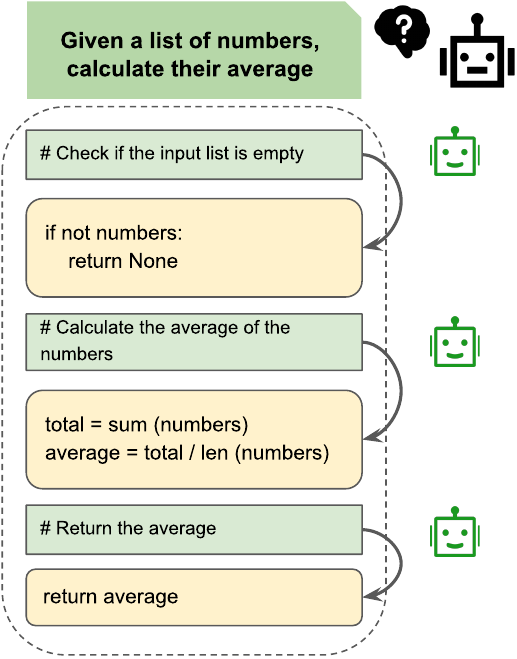}
    \caption{An illustration of the function of code-comment in code LLMs. The comment contributes to breaking down intermediate steps that correspond to the problem description and form an aligned structure with adjacent code lines.}
    \label{fig:overview}
\end{figure}

\section{Introduction}

Recently, techniques on pre-trained Code Large Language Models (Code LLMs) are rapidly developing~\cite{zhao2023survey} and perform well on various code-related tasks~\cite{roziere2023codellama, chen2021evaluating}.
Code generation is to generate code snippets through prompting approaches for the given natural language description and optional extra information ({\em e.g,} data type, function name, and unit test examples). Additionally, the natural language description usually contains complex logic, making direct prompting difficult to solve hard problems.
Decomposing complex problems into easier intermediate steps via the Chain-of-Thought~\cite{wei2022chain} prompting strategy has shown promising performance in general complex tasks. For code generation, the follow-up works divide the planning and code synthesis process into two stages with uni-task few-shot prompts~\cite{jiang2023self-plan} or intermediate steps with tree structure~\cite{zelikman2023parsel}. In addition, constraining the intermediate steps using code structure information can also further improve the coding ability~\cite{li2023structured}.
However, task decomposing and planning are high-level and complex tasks that depend on the most advanced ability of LLMs~\cite{zelikman2023parsel}. 
Consequently, enhancing the code generation of small to medium-sized LLMs efficiently presents a compelling and challenging issue.

To address the above issue, we propose the idea for the first time that code comments serve as the inherent logical pivot bridging natural language and programming language. 
Generally, comments within the code are commonly integral to code corpus. Consequently, during the pre-training stage, training on code corpus endows the pre-trained code models with the respective capacities for understanding and generating code comments. As depicted in Figure~\ref{fig:overview}, code comments can decompose the problem description using natural language, and each comment line establishes an alignment with the neighboring lines of code.
Therefore, we hypothesize that encouraging models to generate comments can easily and effectively bridge the code and complex problem descriptions. 
In order to test our hypothesis, we discuss the possible inserting position of intermediate steps and use cross-entropy loss to evaluate the difficulty of various inserting style codes. According to the statistical results, we found that code in comment style is the easiest for various code models, which provides evidence for the hypothesis.
Based on the above hypothesis, we propose comMents As Natural loGic pivOts (\model) for code generation with problem description decomposition via comments. \model includes a logical comment decoding strategy and comment contrastive learning loss.
Specifically in the training phase, we generate negative samples without comments using the code data with comments in open-source datasets to strengthen the model preference for code with comments. During the decoding stage, the logical comment decoding strategy is adopted to guide the model to explain the code logic via inline comments.

We conduct experiments on HumanEval and MBPP test sets using three backbone models from 3B to 7B.
The experimental result shows that \model improves the pass rate consistently, {\em e.g.,} up to 7.52 pass@10 on HumanEval for StarCoder-7B and up to 4.17 pass@10 on MBPP for WizardCoder-7B.
Our ablation studies show that each component of the method positively contributes to the performance of models, and the method is robust on the hyper-parameter and the various logical comment prompting styles. Furthermore, \model keeps consistent effectiveness in small-size models ({\em e.g., } with pass@10 improvements of up to 3.87 on 3B and 4.07 on 7B), while CoT prompting can lead to severe decline, especially in 3B models. 
The error distribution and code feature data statistics are also provided for fine-grained analysis.

In summary, our contributions are as follows\footnote{We uploaded the code as support material for review and it will be released on Github upon publication.}:
\begin{itemize}
\item We first present that comments are pivots bridging natural language and code language, and conduct an analysis on comparing the difficulty of different positions of inserted decomposition steps for complex problem descriptions. To the best of our knowledge, this is the first work that fully explores the significant advantages of code-comments for coding problem decomposition.
\item We propose \model that includes the contrastive training method and comment prompting strategy. \model improved the code generation ability of the models by strengthening the preference for code with comments and encouraging the model to use comments.
\item The comparison between the CoT prompting strategy and our method indicates the effectiveness of code comments on small and middle model sizes. We conducted robustness evaluations on CoT and logical comment prompting strategies, respectively, and found that LCP achieves a much better performance with smaller standard deviations compared with CoT prompting.
\end{itemize}

\section{Preliminary Analysis: Token-level Loss of Comment Related Code Styles}

\begin{figure}
    \centering
    \includegraphics[width=\linewidth]{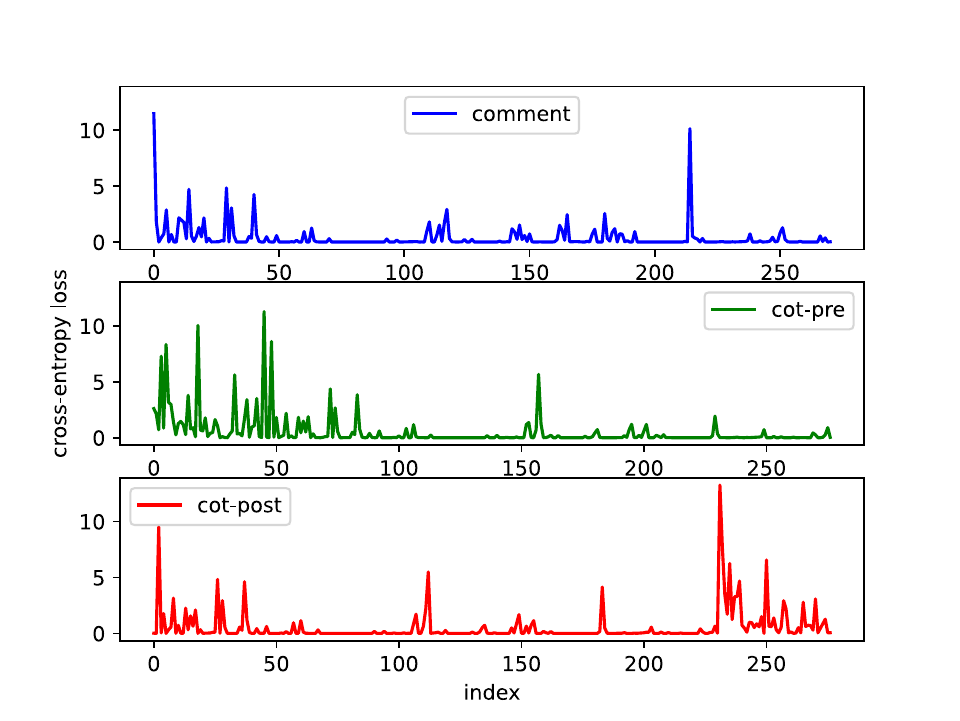}
    \caption{The token-level cross-entropy loss of three code styles for a code problem example in the WizardCoder3B model, and the abscissa represents the token index of the tokenized code.}
    \label{fig:token_loss}
\end{figure}

Braking difficult problems into easier intermediate steps benefits the code generation ability of LLMs in many existing works especially for API-based close-source models. However, most of the task decomposition methods can work only on the most advanced models like GPT-4, and their applicability to open-source models with small or middle sizes is limited. Therefore, we hypothesize that lowering the method difficulty of decomposition methods can benefit lowering the requirement of the model ability, and using comment is one of the possible methods. We define the possible styles for code with comment explanations and quantify their difficulties to models in zero-shot settings.
First, we utilize WizardCoder-3B and WizardCoder-7B as backbone models, and the open-source code generation instruction tuning dataset CodeM as an evaluation dataset. We filtered the data with comments, extracted the comment line, and transferred it to the chain of thoughts.
Then, given a task decomposition step is $T_i$ and the corresponding code block is $C_i$ (where $1\leq i \leq n$, and $n$ is the code block number), we construct three types of code, including the code with the preceding chain of thoughts (CoT-pre, which can be presented as $\{T_1, ..., T_n, C_1, ..., C_n\}$), the following chain of thoughts (CoT-post, which can be presented as $\{C_1, ..., C_n, T_1, ..., T_n\}$), and the inline comments (Comment, which can be presented as $\{T_1, C_1, ..., T_n, C_n\}$).
The transferred CoT-style datasets contain equivalent information compared with the origin comment-style dataset.
Using the token-level cross-entropy loss, we calculate the mean loss of the three datasets in different size models, including 3B and 7B. 
As shown in Figure~\ref{fig:token_loss} comments contain equivalent information with the lowest difficulty. We analyzed the output codes and found that the loss scores of natural language tokens are usually higher than the loss scores of programming language tokens, which is also mentioned in~\citet{zhu2023improving}. However, when the intermediate steps of natural language are inserted in code lines as comments, their loss score decreases obviously. 
In Table~\ref{tab:avg_loss}, we observe that comment loss was significantly much lower than different CoT data, and the gap between cot-style and comment-style is larger when the model size is smaller.
The observations above support our hypothesis that the code containing logical steps as comments lowers decomposition task difficulties for the models (especially for smaller models).

\begin{table}[ht]
    \centering
    \begin{tabular}{lcc}
    \toprule
    Code Type & Wizardcoder 3b & Wizardcoder 7b \\
    \midrule
       CoT-pre  & 0.5318 & 0.4784 \\
       CoT-post  & 0.5606&0.5241 \\
       Comment &0.3384&0.3086 \\

       \bottomrule
    \end{tabular}
    \caption{Average cross-entropy loss for three different code styles: code preceded with step list (CoT-pre), code followed with step list (CoT-post), and inline comments.}
    \label{tab:avg_loss}
\end{table}
\section{\model}
We proposed a simple and effective method \model, including comment contrastive learning loss and logical comment prompt.
\subsection{Backgound: Supervised Fine-tuning on Code Generation Tasks}

The input of a code generation task typically involves a natural language description and an optional programming context. We denote such input as a list of tokens $\mathbf{x}=\{x_1, x_2, \cdots,x_n\}$. Given $x$, the correspondent code snippet $\mathbf{y}=\{y_1, y_2, \cdots, y_m\}$ is expected to be generated by the code LLM $P(\mathbf{y}|\mathbf{x})$.
In standard supervised fine-tuning, the code LLMs are trained to predict the next token based on cross-entropy loss $L_{lm}$ as Equation~\ref{eq:ce_loss}.
\begin{equation}\label{eq:ce_loss}
    L_{lm} = - \sum_{i}^{m} logP(y_i|y_{i-1}, \cdots, y_1, \mathbf{x})
\end{equation}

\subsection{Comment Contrastive Learning}
In code snippets, code comments naturally assume the role of interpreting local code lines using natural language text, where the decomposition of logic also serves as a function of comments. However, directly fine-tuning the loss function treats all tokens as equivalent, without making additional distinctions for comments. To accommodate the task of guiding the model to generate annotations, we adopt a contrastive learning approach to encourage the model to emphasize code comments more during the fine-tuning process.

Our approach consists of the following steps. First, to prepare data for comment contrastive learning, we filter out examples containing comments from the training data. Taking the open-source dataset Code-Python as an example, this dataset contains half of the examples with annotations. For a code snippet $\mathbf{y}_{pos}$ containing comments, we use an open-source code parsing tool \footnote{https://github.com/pygments/pygments} to remove the comments and obtain the non-preferred contrastive sample $\mathbf{y}_{neg}$.
Then, we add a contrastive loss $L_{cl}$ by setting a margin $m$ for the possibility of the labels with comments $\hat{\mathbf{y}}_{pos}$ and without comments $\hat{\mathbf{y}}_{neg}$. 
    \begin{equation}\label{eq:contrast_loss}
    \begin{split}
        L_{cl} = &max(0, m - {\rm log}P_{\theta}(\hat{\mathbf{y}}_{neg}|\mathbf{x})+ \\
        &{\rm log}P_{\theta}(\hat{\mathbf{y}}_{pos}|\mathbf{x})) 
    \end{split}   
    \end{equation}
The final loss $L$ is the addition of the standard cross entropy loss $L_{lm}$ and the contrastive loss $L_{cl}$.
\begin{equation}\label{eq:overall}
    L = L_{lm} + L_{cl}
\end{equation}

\begin{table*}[ht]
    \centering
    \resizebox{0.8\linewidth}{!}{
    \begin{tabular}{ll}
    \toprule
        Prompt Type &  Content \\
        \midrule
        Standard & \begin{tabular}{l}{}
          \#\#\# Instruction:\\Create a Python script for this problem:\\\{input\}\\\#\#\# Response: Here's the Python script for the given problem:
        \end{tabular} \\
        \hdashline
        LCP & \begin{tabular}{l}{}
          \#\#\# Instruction:\\Create a Python script for this problem:\\\{input\}\\\#\#\# Response: \textcolor{blue}{Here's the Python script for the given problem} \\\textcolor{blue}{with comments to explain the logic:}
        \end{tabular} \\
    \bottomrule
    \end{tabular}}
    \caption{Standard instruction-following prompt and LCP (Logical Comment Prompt), and the \textcolor{blue}{blue text} guide the models generate code with comments.}
    \label{tab:commentPrompt}
\end{table*}
\subsection{LCP: Logical Comment Prompt}
In order to enable the model to use comments as intermediate steps intentionally, adding a corresponding instruction in the prompt is the method with minimum cost.
We use the logical comment prompting strategy to guide the model to generate comments explaining the code logic in the decoding stage.
Following the standard prompt including the problem description and instruction for generating code in a certain programming language, the logical comment prompt is shown in Table~\ref{tab:commentPrompt}, which adds \textit{with comments to explain the logic:} or similar text.

\begin{table*}[ht]
    \centering
    \resizebox{0.75\linewidth}{!}{
    \begin{tabular}{lllllll}
    \toprule
    \multirow{2}{*}{setting}   &\multicolumn{3}{c}{HumanEval} & \multicolumn{3}{c}{MBPP}   \\
    \cmidrule(rl){2-4} \cmidrule(rl){5-7}
     &Pass@1&  Pass@5 & Pass@10 & Pass@1&  Pass@5 & Pass@10 \\
    \midrule
    \multicolumn{7}{c}{StarCoder-7B}\\
    SFT  & 35.93 & 55.69 & 63.01 & 42.79 & 56.41 & 61.13 \\
    \w CoT & 22.40 & 49.39 & 58.95 & 26.45 & 52.27 & 59.87 \\
    \w LCP & 39.17 & 61.91 & 70.12 & \textbf{41.11} & \textbf{57.68} & 63.27 \\
    \model & \textbf{39.73} & \textbf{62.81} & \textbf{70.53} & 40.92 & \textbf{57.68} & \textbf{63.53} \\
    \hline
    \multicolumn{7}{c}{WizardCoder-3B}\\
    SFT  &33.37 & 51.43 & 58.12 & 35.98 & 51.51 & 56.80\\
    \w CoT & 17.28 & 41.09 & 51.83 & 21.38 & 44.89 & 52.20 \\
    \w LCP & 34.17 & \textbf{56.07} & \textbf{63.82} & 34.79 & 51.23 & 57.47 \\
    \model &  \textbf{34.47} & 54.55 & 61.99 & \textbf{35.53} & \textbf{52.04} & \textbf{58.27} \\
    \hline
    \multicolumn{7}{c}{WizardCoder-7B}\\
    SFT & 52.80 & 69.74 & 74.59 & 42.73 & 56.77 & 62.33  \\
    \w COT & 42.50 & 68.68 & 75.20 & 31.25 & 52.65 & 59.67 \\
    \w LCP & 52.95 & 70.70 & 75.81 & 41.96 & 57.82 & 63.27 \\
    \model &  \textbf{54.59} & \textbf{72.76} & \textbf{78.66} & \textbf{46.91} & \textbf{62.26} & \textbf{67.87}   \\  
    \bottomrule
    \end{tabular}
    }
    \caption{The main experiments on \model. The \textbf{bold} text is the best result in experiments or comparable settings.}
    \label{tab:main_table}
\end{table*}

\section{Experiments and Results}
In this section, we conduct experiments for our method on different benchmarks and backbone models, and the following sections describe the details of the experiments. 

\subsection{Trainning Settings}
We selected the state of the arts open-source backbone WizardCoder and StarCoder from 3B to 7B. 
We use \textbf{codem-python}~\cite{zan2023codem} as training data, which includes 9600 Python examples distilled from GPT-4 using Evol-instruct~\cite{xu2023wizardlm}. 
Our training script mainly follows WizardCoder~\cite{luo2023wizardcoder} and trains 3 epochs (111 steps) with batch size 256; the warmup step is 15. For the hyperparameter margin $m$ in our comment contrastive training loss, we set $m=0.1$ for all models in the experiments.
We use DeepSpeed\footnote{https://github.com/microsoft/DeepSpeed} Stage 1 for distribution training. Our standard supervised fine-tuning implementation results have an error margin within 1\% the Pass@1 result in the greedy search strategy in CodeM-Python~\cite{zan2023codem}, which indicates that our experiment setting leads to consistent results compared with the existing work.

\paragraph{SFT (Supervised Fine-Tuning)} We construct the standard instruction template in WizardCoder~\cite{luo2023wizardcoder} for the CodeM-Python dataset, and the training objection is cross entropy. 
\paragraph{CoT (Chain-of-Thought)} As another baseline, this setting uses zero-shot-CoT~\cite{kojima2022large} on the standardly fine-tuned models, and the detailed prompt is the ``CoT-pre1'' appended in the Appendix~\ref{sec:prompt} Table~\ref{tab:seed_prompts}. 
\paragraph{LCP (Logical Comment Prompt)} This setting uses the LCP strategy to guide standardly fine-tuned models generating codes with comments. The detail prompt used in experiments is shown in Table~\ref{tab:commentPrompt}.
\paragraph{\model} (comMents As Natural loGic pivOts) During the training stage, \model trains the same dataset in the same instruction template, and the training objection is comment contrastive loss following Equation~\ref{eq:contrast_loss} and Equation~\ref{eq:overall}.
During the inference stage, the logical comment prompt (LCP) strategy will be used.
\paragraph{Decoding Settings}
We follow the setting in CodeLLaMA~\cite{roziere2023codellama}, using temperature $T = 0.8$  and nuclear sampling $top\_p=0.95$. 
We generate n = 10 samples and calculate Pass@k, where k = 1, 5, 10.

\subsection{Evaluation Settings}
\paragraph{Evaluate Metrics}  Pass@k~\cite{chen2021evaluating} is currently the most widely used metric in code generation. It means with sampling $n$ samples, the possibility of any code is correct in $k$ samples.
\begin{equation}
    {\rm Pass} @ k=\mathbb{E}_{\text{Problems}}\left[1-\frac{\left(\begin{array}{c}n-c \\ k\end{array}\right)}{\left(\begin{array}{l}n \\ k\end{array}\right)}\right]
\end{equation}
\paragraph{HumanEval} HumanEval~\cite{chen2021evaluating} is currently the most widely used dataset for code generation. It contains 164 problems written in Python that evaluate programming language comprehension, algorithms, and basic mathematics. For each problem, 7.7 unit tests are contained on average. 
\paragraph{MBPP} MBPP~\cite{austin2021mbpp} test set contains 500 Python problems. The problems cover basic programming concepts and standard library functionality. Each problem comprises a task description, code solution, and 3 automated test cases.

\subsection{Main Result}
In assessing the efficacy of the \model, we utilize pass rates on HumanEval and MBPP as our primary evaluative measure. The principal results are presented in Table~\ref{tab:main_table}. Overall, there are three main aspects that we observed.\\
First, comparing the SFT baseline with the \model, consistent enhancements are demonstrated across various sizes and backbones. Notably, the StarCoder-7B exhibits a significant augmentation, with an increase of 7.52 percentage points for HumanEval in Pass@10. In the case of WizardCoder-7B, there is a 4.17 percentage points enhancement in the Pass@10 outcome for the MBPP test set. 
Secondly, comparing the pass rates of the backbones in the CoT settings, we observe a decline in model performance in all instances, with a more pronounced gap evident in the smaller 3B model. In contrast, LCP prompting consistently contributes to the enhancement of model performance, demonstrating its stability and effectiveness.
Thirdly, in the ablation of LCP and \model, we observed that on the HumanEval and MBPP test set, a standardly finetuned StarCoder-7B can achieve a significantly higher score simply by prompting with LCP. This observation underscores the substantial potential of low-cost prompting methods.

Considering the convenience of providing comparable results on zero-shot settings, we also provide the greedy search results of Pass@1 for comparison with the results of existing works. Since Pass@1 is the most strict indicator, it is more difficult to improve than Pass@10. According to results in Table~\ref{tab:pass1}, the zero-shot CoT prompting persistently results in diminished model performance; however, an enhancement in Pass@1 is observable when employing LCP prompting.
\begin{table}[ht]
    \centering
    \resizebox{\linewidth}{!}{
    \begin{tabular}{lccc}
    \toprule
       Model  &  Pass@1 & Pass@5 & Pass@10\\
       \midrule

        GPT-4 & 67.00* & -- & --\\

        CodeLLaMA-7B & 38.40* & -- & --\\

     StarCoder-7B & 26.83* & -- &--\\

       WizardCoder-7B & 55.50* &76.03& 82.31\\
       \hline
       WizardCoder-3B & 34.80* & 51.72& 59.75 \\

        w/ CoT& 25.61 & 40.05 & 48.78\\
        w/ LCP & 36.59  & 51.57& 59.15\\

         \bottomrule
        
    \end{tabular}
    }
    \caption{Zero-shot performance of various backbones on HumanEval. The Pass@1 is in greedy search decoding and the Pass@5/10 is in sampling decoding with T=0.8. We use * to denote the results from~\citet{achiam2023gpt}, ~\citet{roziere2023codellama},~\citet{li2023starcoder} and~\citet{luo2023wizardcoder}.}
    %
    \label{tab:pass1}
\end{table}

\subsection{Ablation Study on the Training Method}
We evaluate the efficacy of the two components of \model through an ablation study conducted on WizardCoder-7B. The test set employed is HumanEval, and we adhere to the main experimental settings with T=0.8 and p=0.95. Based on the results presented in Table \ref{tab:ablation} and Table~\ref{tab:main_table}, both the comment prompt and the contrastive training loss contribute to an increase in the Pass@10 passing rate. Furthermore, integrating these two components can further augment the performance in terms of the passing rate.
\begin{table}[ht]
    \centering
    \resizebox{0.8\linewidth}{!}{
    \begin{tabular}{lccc}
    \toprule
       Setting  & Pass@1 & Pass@5 & Pass@10 \\
    \midrule
        SFT &52.80&	69.74&	74.59 \\
        w/ LCP &52.95&	70.70&	75.81 \\
        w/ $L_{cl}$ &\textbf{55.47}&	71.75&	76.02\\
        \model &54.59&	\textbf{72.76}&	\textbf{78.66}\\
    \bottomrule
    \end{tabular}
    }
    \caption{Ablation study on contrastive loss of \model based on WizardCode-7B. }
    \label{tab:ablation}
\end{table}

\subsection{The Effect of Margin}\label{sec:margin}
We marginalize the representation of code without comments with code LLMs, and the hyper-parameter margin $m$ is used to control the extent of negative marginalization.
We examine various margin settings for WizardCoder-7B on HumanEval and follow decoding setting T=0.8. The results in Table~\ref{tab:margin_abl} indicate that \model can outperform the baselines under margins larger than 0.05. However, when the margin is 0.03, the performance of the model decreases significantly both on Pass@5 and Pass@10, which indicates the model is sensitive to the margin value.
\begin{table}[ht]
    \centering
    \resizebox{0.8\linewidth}{!}{
    \begin{tabular}{lccc}
       \toprule
    Margin  & Pass@1 & Pass@5 & Pass@10 \\
     \midrule 
     0.03 & 53.11&	70.42&	75.20\\
     0.05 & 53.66&	72.98&	78.66\\
     0.10 & 54.59&	72.76&	78.66\\
     0.15 & 53.13&	73.00&	79.88\\
     \bottomrule
    \end{tabular}}
    \caption{Roubustness validation on margin.}
    \label{tab:margin_abl}
\end{table}

\section{Analysis}
\subsection{The Robustness Study Against LCP and CoT}\label{sec:prompt}

We rephrase the prompt into several versions to mitigate the model's randomness in response to different prompt styles. Initially, we manually created three prompts for LCP and CoT, which served as seed prompts. Subsequently, we employ GPT-4 to generate four additional variants for each prompt. Furthermore, for a more nuanced analysis, we segregate CoT into two categories: the first involves specifying the chain of thought text before generating the final code, representing the most typical and universal form; and the second is a freestyle CoT devoid of position-constraining guidance. 
Ultimately, we have 15 prompts for each category. The complete prompts are provided in the Appendix~\ref{sec:var_prompt}.

The result in Table~\ref{tab:prompt_mean_var} demonstrated that the LCP prompting strategy yields a considerably higher average performance and a lower standard deviation. It is worth noting that the CoT without position constraint has higher performance than typical CoT on average with a larger deviation. 

Apart from the most commonly used independent natural language plan before actual coding, the model can also interpret CoT as explanatory comments for the code. The LCP method proposes that intermediate steps should be incorporated as comments, thereby stabilizing the behavior of models and enhancing their performance compared to generating code problems directly.
\begin{table}[ht]
    \centering
    \resizebox{0.85\linewidth}{!}{
    \begin{tabular}{lccc}
    \toprule
    Setting & Pass@1 & Pass@5 & Pass@10 \\
    \midrule
    \multicolumn{4}{c}{CoT-Pre} \\
      Avg   & 33.24 & 61.88 & 71.38 \\
      StDEV &9.21  & 6.45  & 5.43 \\
      \hdashline
      \multicolumn{4}{c}{LCP} \\
      Avg  &48.65 & 68.85 & 74.76\\
      StDEV &1.37  & 1.74  & 2.45  \\
      \hdashline
      \multicolumn{4}{c}{CoT-No-Position} \\
      Avg   & 40.69 & 65.05 & 72.03  \\
      StDEV &10.64 & 7.19  & 5.46  \\
      \bottomrule
    \end{tabular}}
    \caption{The mean and standard deviation of three prompt groups. CoT-Pre means the prompts request models that generate the Chain-of-Thought first and then generate code, while CoT-No-Position means the prompts guide models generate code with Chain-of-Thought only. }
    \label{tab:prompt_mean_var}
\end{table}

\subsection{Statistical Features of Generated Codes}\label{sec:data_stat}
We statistic the style transformation between the different instruction strategies and training methods, including the effective code line number and comment line number. As Tabel~\ref{tab:style_count}, the CoT prompt guides model generates intermedia steps before generating code, leading to fewer comment lines than the prompts without guidance. The LCP prompt induces more code comments compared with the origin prompt. Finally, the contrastive training loss $L_{cl}$ does not change the code and comment line number features.

\begin{table}[ht]
    \centering
    \resizebox{0.95\linewidth}{!}{
    \begin{tabular}{lcc}
    \toprule
        setting & \#avg. code line & \#avg. comment line  \\
    \midrule
       origin  & 15.56 & 1.36\\
       \w CoT & 11.68 & 0.38 \\
       \w LCP  & 20.92 & 3.29\\
       \model &20.80 & 3.33 \\
    \bottomrule
    \end{tabular}
    }
    \caption{Code and comment line counting on different prompts in the SFT setting and \model results based on WizardCoder-7B.}
    \label{tab:style_count}
\end{table}

\subsection{The Relationship Between CoT and Contrastive Comment Loss}

To investigate whether the enhancement of comparative learning for comments can be directly generalized to the capability of CoT. Given the substantial fluctuation in the code generated by CoT's prompt, we selected three typical prompts as the subjects of study. Through the statistical analysis of the number of lines in the generated code, these three prompts respectively represented code styles with three different mean values of 0, 2, and 3 on WizardCoder-7B, and the detailed prompt will be listed in the Appendix (CoT1: CoT-No-Position1, CoT2: CoT-pre1, CoT3: CoT-pre2). 
We test the model in the SFT setting and the contrastive trained setting. 
The results are shown in Table.~\ref{tab:cot_comment}. Comparing the mean of comment lines and the final pass rates, we can find that the more comments the CoT prompt can guide, the higher performance can be seen. 

Comparing SFT and $L_{cl}$ under the same settings, it can be observed that for the prompt that inherently performs CoT through code comments (i.e., ``CoT1''), the models after contrastive learning consistently improve code pass rates. However, for prompts that perform CoT through other means, a decline in performance is observed in the 3B models after comparative learning. This phenomenon suggests that merely fine-tuning through a small amount of contrastive learning on comments is not easily transferable to other forms of task decomposition.
\begin{table}[ht]
    \centering
    \resizebox{\linewidth}{!}{
    \begin{tabular}{llllll}
    \toprule
\multicolumn{2}{c}{setting}& pass@1 & pass@5 & pass@10 & MCL \\
\midrule
\multicolumn{6}{c}{WizardCoder-3B}   \\
\multirow{2}{*}{CoT1}    & SFT       & 28.80 & 52.22 & 61.38 & 2.83  \\
        & $L_{cl}$ & 30.10 & 52.85 & 60.77 & 2.64  \\
\multirow{2}{*}{CoT2}     & SFT       & 17.91 & 44.05 & 53.86 & 1.35  \\
        & $L_{cl}$ & 15.65 & 40.84 & 52.03 & 1.06  \\
\multirow{2}{*}{CoT3}    & SFT       & 17.28 & 41.09 & 51.83 & 0.95  \\
        & $L_{cl}$ & 12.85 & 34.97 & 45.33 & 1.03  \\
        \hline
\multicolumn{6}{c}{WizardCoder-7B}   \\
\multirow{2}{*}{CoT1}    & SFT       & 47.64 & 70.89 & 77.64 & 2.70  \\
        & $L_{cl}$ & 50.43 & 71.68 & 77.64 & 3.03  \\
\multirow{2}{*}{CoT2}     & SFT       & 47.70 & 67.94 & 73.78 & 1.86  \\
        & $L_{cl}$ & 44.80 & 70.28 & 76.42 & 1.89  \\
\multirow{2}{*}{CoT3}     & SFT       & 42.30 & 67.76 & 74.19 & 0.38  \\
        & $L_{cl}$ & 41.02 & 68.31 & 75.81 & 0.26  \\
        \bottomrule
\end{tabular}}
    \caption{The relevance of the CoT prompting performance and the training loss setting. ``MCL'' is the abbreviation of mean comment lines. }
    \label{tab:cot_comment}
\end{table}

\subsection{Error Distribution}\label{sec:error}

To analyze the pass rate improvement of \model, we classify problem description understanding error types using the HumanEval test set. \model improves model output by utilizing code comments, which does not target a specific type of code error but rather improves the understanding of problem descriptions from the overall perspective via decomposing complex coding logic. Meanwhile, in the code generation task, test cases can effectively reflect the degree of understanding of the output results.
We established two thresholds for evaluating problem understanding: whether the compiler can successfully compile the code and whether the code can pass at least one test case. Based on this, we roughly classified the understanding level into three types from low to high.

We divide the failed cases into three categories: \textbf{Verification Error (VE)} includes Runtime Error, Syntax Error, and the other case except for Wrong Answer; \textbf{All Wrong Answer (AWA)} includes the cases that cannot pass the first test cases of the scripts, indicating that the generated code has a poor understanding on problem descriptions; \textbf{Particle Wrong Answer (PWA)} includes the cases that can pass the first unit test at least but fail finally.

\begin{table}[ht]
    \centering
    \resizebox{0.9\linewidth}{!}{
    \begin{tabular}{lcccc}
    \toprule
       Setting  & VE & AWA & PWA & overall \\
    \midrule
        SFT & 4.31& 10.24& 11.87 & 26.42 \\
        w/ LCP & 3.84& 8.88 & 11.46 & 24.19 \\
        w/ $L_{cl}$ & \textbf{2.81} & 10.69 &	10.49&23.99 \\
        \model&2.84&\textbf{8.49}&\textbf{10.00} &\textbf{21.33}\\
    \bottomrule
    \end{tabular}}
    \caption{Test result error distribution statistics.}
    \label{tab:error_dist}
\end{table}

As shown in Tabel~\ref{tab:error_dist}, compared with the error rates in SFT, \model decreases all three types of error rates. Furthermore, the ablation components also show lower error rates in three types, indicating that \model has reduced the code error rate across various degrees of comprehension error.

\section{Related Work}
\subsection{Code LLMs and Logical Reasoning}
The rapid development of pre-trained language models brings the bloom of code LLMs. The most advanced open-source code LLMs are pre-trained by a large amount of natural language corpus and code corpus~\cite{roziere2023codellama}. Based on the backbones, WizardCoder~\cite{luo2023wizardcoder} and OctoPack~\cite{muennighoff2023octopack} proposed instruction-tuning data construction methods that boost the instruction-following ability of code LLMs significantly. Besides, ~\citet{fu2022does} proposed that code corpus can be the source of CoT ability. Both~\citet{fu2022does} and~\citet{ma2023training} observed that training small-size code LLMs on code data can augment the logical reasoning ability of the model further.

\subsection{Prompting-based Code Generation Strategy}
The planning strategy is the mainstream technique to transform the causal language model to fit logical text generation.
Existing works include the variant of CoT (Chain of Thought)~\cite{jiang2023self-plan, li2023structured}, Reflexion~\cite{shinn2023reflexion}, etc.
Most prompting strategies utilize the outstanding logical and understanding abilities of the most advanced models. However, these strategies will not work even on large open-source models with a size beyond 10B~\cite{jiang2023self-plan,shinn2023reflexion}.
Another mainstream method is compiling feedback to train code models~\cite{chen2023improving, chen2022codet}.
Our method utilizes the comment, the widely accepted rule for coders, to enhance the code readability and explain the code logic simultaneously. ~\citet{li2023think} proposed that the brainstorming step can elite code generation performance. The commonness of comments in training code corpus enables models to easily capture the relationship between code and natural language description logic in different sizes of backbone models.



\section{Conclusion}
In this paper, we first emphasize the importance of code comments in bridging natural language and code language, and then we hypothesize that more comments to explain code logic can enhance code generation performance. We propose a simple and effective method \model, which contains contrastive comment loss and logical comment prompts. The main experiments and ablation studies show that \model effectively augments the pass rate of code generated by the models. In further analysis, we also compared the LCP and CoT prompting strategies and found that LCP has a significantly better and stabler performance comparing CoT, especially for smaller models. The comment contrastive training can also boost CoT performance in certain conditions.\par
In conclusion, our work introduces the comment pivot as a novel perspective, and guiding models to use comments can augment the code generation ability stably.
In the future, we will further explore the application potential of comments in broader complex scenarios, such as tool usage and LLM agents.


\bibliography{acl_latex}

\begin{thebibliography}{22}
\expandafter\ifx\csname natexlab\endcsname\relax\def\natexlab#1{#1}\fi

\bibitem[{Achiam et~al.(2023)Achiam, Adler, Agarwal, Ahmad, Akkaya, Aleman,
  Almeida, Altenschmidt, Altman, Anadkat et~al.}]{achiam2023gpt}
Josh Achiam, Steven Adler, Sandhini Agarwal, Lama Ahmad, Ilge Akkaya,
  Florencia~Leoni Aleman, Diogo Almeida, Janko Altenschmidt, Sam Altman,
  Shyamal Anadkat, et~al. 2023.
\newblock Gpt-4 technical report.
\newblock \emph{arXiv preprint arXiv:2303.08774}.

\bibitem[{Austin et~al.(2021)Austin, Odena, Nye, Bosma, Michalewski, Dohan,
  Jiang, Cai, Terry, Le et~al.}]{austin2021mbpp}
Jacob Austin, Augustus Odena, Maxwell Nye, Maarten Bosma, Henryk Michalewski,
  David Dohan, Ellen Jiang, Carrie Cai, Michael Terry, Quoc Le, et~al. 2021.
\newblock Program synthesis with large language models.
\newblock \emph{arXiv preprint arXiv:2108.07732}.

\bibitem[{Chen et~al.(2023)Chen, Scheurer, Korbak, Campos, Chan, Bowman, Cho,
  and Perez}]{chen2023improving}
Angelica Chen, J{\'e}r{\'e}my Scheurer, Tomasz Korbak, Jon~Ander Campos,
  Jun~Shern Chan, Samuel~R Bowman, Kyunghyun Cho, and Ethan Perez. 2023.
\newblock Improving code generation by training with natural language feedback.
\newblock \emph{arXiv preprint arXiv:2303.16749}.

\bibitem[{Chen et~al.(2022)Chen, Zhang, Nguyen, Zan, Lin, Lou, and
  Chen}]{chen2022codet}
Bei Chen, Fengji Zhang, Anh Nguyen, Daoguang Zan, Zeqi Lin, Jian-Guang Lou, and
  Weizhu Chen. 2022.
\newblock Codet: Code generation with generated tests.
\newblock In \emph{The Eleventh International Conference on Learning
  Representations}.

\bibitem[{Chen et~al.(2021)Chen, Tworek, Jun, Yuan, Pinto, Kaplan, Edwards,
  Burda, Joseph, Brockman et~al.}]{chen2021evaluating}
Mark Chen, Jerry Tworek, Heewoo Jun, Qiming Yuan, Henrique Ponde de~Oliveira
  Pinto, Jared Kaplan, Harri Edwards, Yuri Burda, Nicholas Joseph, Greg
  Brockman, et~al. 2021.
\newblock Evaluating large language models trained on code.
\newblock \emph{arXiv preprint arXiv:2107.03374}.

\bibitem[{Fu et~al.(2022)Fu, Peng, and Khot}]{fu2022does}
Yao Fu, Hao Peng, and Tushar Khot. 2022.
\newblock How does gpt obtain its ability? tracing emergent abilities of
  language models to their sources.
\newblock \emph{Yao Fu’s Notion}.

\bibitem[{Jiang et~al.(2023)Jiang, Dong, Wang, Shang, and
  Li}]{jiang2023self-plan}
Xue Jiang, Yihong Dong, Lecheng Wang, Qiwei Shang, and Ge~Li. 2023.
\newblock Self-planning code generation with large language model.
\newblock \emph{arXiv preprint arXiv:2303.06689}.

\bibitem[{Kojima et~al.(2022)Kojima, Gu, Reid, Matsuo, and
  Iwasawa}]{kojima2022large}
Takeshi Kojima, Shixiang~Shane Gu, Machel Reid, Yutaka Matsuo, and Yusuke
  Iwasawa. 2022.
\newblock Large language models are zero-shot reasoners.
\newblock \emph{Advances in neural information processing systems},
  35:22199--22213.

\bibitem[{Li et~al.(2023{\natexlab{a}})Li, Li, Li, and Jin}]{li2023structured}
Jia Li, Ge~Li, Yongmin Li, and Zhi Jin. 2023{\natexlab{a}}.
\newblock Structured chain-of-thought prompting for code generation.
\newblock \emph{arXiv preprint arXiv:2305.06599}.

\bibitem[{Li et~al.(2023{\natexlab{b}})Li, Allal, Zi, Muennighoff, Kocetkov,
  Mou, Marone, Akiki, Li, Chim et~al.}]{li2023starcoder}
Raymond Li, Loubna~Ben Allal, Yangtian Zi, Niklas Muennighoff, Denis Kocetkov,
  Chenghao Mou, Marc Marone, Christopher Akiki, Jia Li, Jenny Chim, et~al.
  2023{\natexlab{b}}.
\newblock Starcoder: may the source be with you!
\newblock \emph{arXiv preprint arXiv:2305.06161}.

\bibitem[{Li et~al.(2023{\natexlab{c}})Li, Xue, Xie, and Li}]{li2023think}
Xin-Ye Li, Jiang-Tian Xue, Zheng Xie, and Ming Li. 2023{\natexlab{c}}.
\newblock Think outside the code: Brainstorming boosts large language models in
  code generation.
\newblock \emph{arXiv preprint arXiv:2305.10679}.

\bibitem[{Luo et~al.(2023)Luo, Xu, Zhao, Sun, Geng, Hu, Tao, Ma, Lin, and
  Jiang}]{luo2023wizardcoder}
Ziyang Luo, Can Xu, Pu~Zhao, Qingfeng Sun, Xiubo Geng, Wenxiang Hu, Chongyang
  Tao, Jing Ma, Qingwei Lin, and Daxin Jiang. 2023.
\newblock Wizardcoder: Empowering code large language models with
  evol-instruct.
\newblock \emph{arXiv preprint arXiv:2306.08568}.

\bibitem[{Ma et~al.(2023)Ma, Liu, Yu, Zhang, Jiang, Wang, and
  Li}]{ma2023training}
Yingwei Ma, Yue Liu, Yue Yu, Yuanliang Zhang, Yu~Jiang, Changjian Wang, and
  Shanshan Li. 2023.
\newblock At which training stage does code data help llms reasoning?
\newblock \emph{arXiv preprint arXiv:2309.16298}.

\bibitem[{Muennighoff et~al.(2023)Muennighoff, Liu, Zebaze, Zheng, Hui, Zhuo,
  Singh, Tang, von Werra, and Longpre}]{muennighoff2023octopack}
Niklas Muennighoff, Qian Liu, Armel Zebaze, Qinkai Zheng, Binyuan Hui,
  Terry~Yue Zhuo, Swayam Singh, Xiangru Tang, Leandro von Werra, and Shayne
  Longpre. 2023.
\newblock Octopack: Instruction tuning code large language models.
\newblock \emph{arXiv preprint arXiv:2308.07124}.

\bibitem[{Roziere et~al.(2023)Roziere, Gehring, Gloeckle, Sootla, Gat, Tan,
  Adi, Liu, Remez, Rapin et~al.}]{roziere2023codellama}
Baptiste Roziere, Jonas Gehring, Fabian Gloeckle, Sten Sootla, Itai Gat,
  Xiaoqing~Ellen Tan, Yossi Adi, Jingyu Liu, Tal Remez, J{\'e}r{\'e}my Rapin,
  et~al. 2023.
\newblock Code llama: Open foundation models for code.
\newblock \emph{arXiv preprint arXiv:2308.12950}.

\bibitem[{Shinn et~al.(2023)Shinn, Cassano, Gopinath, Narasimhan, and
  Yao}]{shinn2023reflexion}
Noah Shinn, Federico Cassano, Ashwin Gopinath, Karthik~R Narasimhan, and Shunyu
  Yao. 2023.
\newblock Reflexion: Language agents with verbal reinforcement learning.
\newblock In \emph{Thirty-seventh Conference on Neural Information Processing
  Systems}.

\bibitem[{Wei et~al.(2022)Wei, Wang, Schuurmans, Bosma, Xia, Chi, Le, Zhou
  et~al.}]{wei2022chain}
Jason Wei, Xuezhi Wang, Dale Schuurmans, Maarten Bosma, Fei Xia, Ed~Chi, Quoc~V
  Le, Denny Zhou, et~al. 2022.
\newblock Chain-of-thought prompting elicits reasoning in large language
  models.
\newblock \emph{Advances in Neural Information Processing Systems},
  35:24824--24837.

\bibitem[{Xu et~al.(2023)Xu, Sun, Zheng, Geng, Zhao, Feng, Tao, and
  Jiang}]{xu2023wizardlm}
Can Xu, Qingfeng Sun, Kai Zheng, Xiubo Geng, Pu~Zhao, Jiazhan Feng, Chongyang
  Tao, and Daxin Jiang. 2023.
\newblock Wizardlm: Empowering large language models to follow complex
  instructions.
\newblock \emph{arXiv preprint arXiv:2304.12244}.

\bibitem[{Zan et~al.(2023)Zan, Yu, Shen, Zhang, Chen, Geng, Chen, Ji, Yao, Wang
  et~al.}]{zan2023codem}
Daoguang Zan, Ailun Yu, Bo~Shen, Jiaxin Zhang, Taihong Chen, Bing Geng, Bei
  Chen, Jichuan Ji, Yafen Yao, Yongji Wang, et~al. 2023.
\newblock Can programming languages boost each other via instruction tuning?
\newblock \emph{arXiv preprint arXiv:2308.16824}.

\bibitem[{Zelikman et~al.(2023)Zelikman, Huang, Poesia, Goodman, and
  Haber}]{zelikman2023parsel}
Eric Zelikman, Qian Huang, Gabriel Poesia, Noah Goodman, and Nick Haber. 2023.
\newblock Parsel: Algorithmic reasoning with language models by composing
  decompositions.
\newblock In \emph{Thirty-seventh Conference on Neural Information Processing
  Systems}.

\bibitem[{Zhao et~al.(2023)Zhao, Zhou, Li, Tang, Wang, Hou, Min, Zhang, Zhang,
  Dong et~al.}]{zhao2023survey}
Wayne~Xin Zhao, Kun Zhou, Junyi Li, Tianyi Tang, Xiaolei Wang, Yupeng Hou,
  Yingqian Min, Beichen Zhang, Junjie Zhang, Zican Dong, et~al. 2023.
\newblock A survey of large language models.
\newblock \emph{arXiv preprint arXiv:2303.18223}.

\bibitem[{Zhu et~al.(2023)Zhu, Li, Li, Zhao, Li, Jin, and
  Mei}]{zhu2023improving}
Yuqi Zhu, Jia~Allen Li, Ge~Li, YunFei Zhao, Jia Li, Zhi Jin, and Hong Mei.
  2023.
\newblock Improving code generation by dynamic temperature sampling.
\newblock \emph{arXiv preprint arXiv:2309.02772}.

\end{thebibliography}

\appendix

\section{The Detail of Variant Prompts}\label{sec:var_prompt}
We use three types of prompts to generate codes with different styles. The CoT-pre indicates prompts that request models to generate a step-by-step chain of thoughts before generating code, and the CoT-No-Position indicates prompts that only request models generate a step-by-step chain of thoughts.
The concrete prompts of three types are shown as follows:
We show the concrete prompts of the three seed prompts for each type in Table~\ref{tab:seed_prompts}, and the other prompts are paraphrased by GPT-4.
\begin{table*}[h]
    \centering
    \begin{tabular}{lp{12cm}}
    \toprule
    prompt name & content \\
    \midrule
    CoT-pre1& First, consider the process step by step, and then present the Python script: \\
CoT-pre2& Start by creating a step-by-step coding plan, then follow it with the Python script for the provided issue: \\
CoT-pre3& The following Python script addresses the specified issue and includes a detailed explanation at the start: \\
\hdashline  
Comment1& Here is the annotated Python script that solves the specified issue: \\
Comment2& To clarify the logic, it's preferable to include annotations in the script. Here's the Python code for the given issue: \\
Comment3& The following Python script addresses the given problem, with complex logic broken down using straightforward comments:\\
\hdashline
CoT-No-Position1 & Here's the Python code for the specified issue, broken down into logical steps: \\
CoT-No-Position6 & The Python program for the aforementioned issue is presented here, with a step-by-step explanation: \\
CoT-No-Position11& Here's a Python script that tackles the given problem, accompanied by a comprehensive breakdown: \\
\bottomrule
    \end{tabular}
    \caption{Three handcraft seed prompts for each type.}
    \label{tab:seed_prompts}
\end{table*}

\section{Case Study}
We use a simple coding task to illustrate the difference between the origin prompt, the LCP prompt, and the CoT prompt. The problem description is shown as follows.
\begin{lstlisting}

def is_simple_power(x, n):
    """Your task is to write a function that returns true if a number x is a simple
    power of n and false in other cases.
    x is a simple power of n if n**int=x
    For example:
    is_simple_power(1, 4) => true
    is_simple_power(2, 2) => true
    is_simple_power(8, 2) => true
    is_simple_power(3, 2) => false
    is_simple_power(3, 1) => false
    is_simple_power(5, 3) => false
    """
\end{lstlisting}

\paragraph{Origin Prompt result \textcolor{red}{[Failed]}} The origin prompt contains no extra instructions, and the generated code fails to pass all of the unit tests because it can only solve the cases provided as examples.
\begin{lstlisting}
```python
def is_simple_power(x, n):
    """Your task is to write a function that returns true if a number x is a simple
    power of n and false in other cases.
    x is a simple power of n if n**int=x
    For example:
    is_simple_power(1, 4) => true
    is_simple_power(2, 2) => true
    is_simple_power(8, 2) => true
    is_simple_power(3, 2) => false
    is_simple_power(3, 1) => false
    is_simple_power(5, 3) => false
    """
    if n == 1:
        return True
    elif x == n:
        return True
    elif x == n**2:
        return True
    elif x == n**3:
        return True
    else:
        return False
```

In this script, we define a function called `is_simple_power` that takes two parameters `x` and `n`. We check a few possible cases to determine if `x` is a simple power of `n`. If any of these cases are true, we return `True`, otherwise we return `False`
\end{lstlisting}

\paragraph{CoT result \textcolor{red}{[Failed]}} The code generated in the CoT prompt setting contains a step-by-step explanation that simply decomposes the problem description into two bullet points. The final generated code fails due to the generated ambiguous reinterpreted bullet points, and there is a large gap between the code implementation and the problem description.
\begin{lstlisting}
    

1. Define a function called `is_simple_power` that takes two parameters, `x` and `n`.
2. Inside the function, check if `x` is equal to `n**int(n)`. If it is, return `True`. Otherwise, return `False`.

Here is the Python script implementing the above steps:

```python
def is_simple_power(x, n):
    if x == n**int(n):
        return True
    else:
        return False
```

This function checks if a number `x` is a simple power of `n`. If `x` is equal to `n` raised to the power of `int(n)`, it returns `True`. Otherwise, it returns `False`.
\end{lstlisting}

\paragraph{LCP result \textcolor{green}{[Passed]}} The code generated in the LCP prompt setting contains clear comments to explain the code logic and all of the unit tests are passed successfully.
\begin{lstlisting}
```python
def is_simple_power(x, n):
    """
    Your task is to write a function that returns true if a number x is a simple
    power of n and false in other cases.
    x is a simple power of n if n**int=x
    For example:
    is_simple_power(1, 4) => true
    is_simple_power(2, 2) => true
    is_simple_power(8, 2) => true
    is_simple_power(3, 2) => false
    is_simple_power(3, 1) => false
    is_simple_power(5, 3) => false
    """

    # Check if x is a perfect power of n
    if n == 1:
        return x == 1

    # Check if x is a simple power of n
    if x % n == 0:
        return is_simple_power(x // n, n)

    # Check if x is a perfect power of n
    for i in range(2, x):
        if x == n ** i:
            return True

    # If x does not satisfy any of the above conditions, it is not a simple power of n
    return False
```

Now, you can call the function `is_simple_power` with the desired inputs to check if a number is a simple power of a given base.
}
\end{lstlisting}

\end{document}